\pdfoutput=1

\documentclass[11pt]{article}

\usepackage{ACL2023}

\usepackage{times}
\usepackage{latexsym}

\usepackage{amsmath, amssymb, amsthm}
\usepackage{graphics, graphicx, bm, physics}
\usepackage{algpseudocode}
\usepackage{algorithm}
\usepackage{rotating}
\usepackage{nicefrac}
\usepackage{subfigure}
\newcommand{\argmin}{\mathop{\mathrm{arg\,min}}}

\usepackage[T1]{fontenc}

\usepackage[utf8]{inputenc}

\usepackage{microtype}

\usepackage{inconsolata}

\usepackage{soul}

\title{Class based Influence Functions for Error Detection}

\author{Nguyen Duc-Thang$^*\dagger$\ \ Hoang Thanh-Tung$^*\dagger$\ \ Quan Tran\thanks{\ \ Joint first authors} \,$\ddagger$ \\
\bf{Huu-Tien Dang$\dagger$\ \ Nguyen Ngoc-Hieu$\dagger$\ \ Anh Dau$\dagger$\ \ Nghi Bui$\dagger$}\\
  $\dagger$ FPT Software AI Center \quad $\ddagger$ Adobe Research \\
  \texttt{\{nguyenducthang8a2, htt210, quanthdhcn\}@gmail.com}
  }

\begin{document}
\maketitle
\begin{abstract}
Influence functions (IFs) are a powerful tool for detecting anomalous examples in large scale datasets. 
However, they are unstable when applied to deep networks.
In this paper, we provide an explanation for the instability of IFs and develop a solution to this problem.
We show that IFs are unreliable when the two data points belong to two different classes.
Our solution leverages class information to improve the stability of IFs.
Extensive experiments show that our modification significantly improves the performance and stability of IFs while incurring no additional computational cost.
\end{abstract}

\section{Introduction}
\label{sec:intro}
Deep learning models are data hungry.
Large models such as transformers \cite{transformer}, BERT \cite{devlin2018bert}, and GPT-3 \cite{gpt3} require millions to billions of training data points.
However, data labeling is an expensive, time consuming, and error prone process.
Popular datasets such as the ImageNet \cite{deng2009imagenet} contain a significant amount of errors - data points with incorrect or ambiguous labels \cite{beyer2020are}.
The need for automatic error detection tools is increasing as the sizes of modern datasets grow.

Influence function (IF) \cite{koh2017understanding} and its variants \cite{charpiat2019input, khanna2019interpreting, barshan2020relatif, pruthi2020estimating} are a powerful tool for estimating the influence of a data point on another data point.
Researchers leveraged this capability of IFs to design or detect adversarial \cite{cohen2020detecting}, poisonous \cite{koh2022stronger, koh2017understanding}, and erroneous \cite{dau2022towards} examples in large scale datasets.
The intuition is that these harmful data points usually have a negative influence on other data points and this influence can be estimated with IFs.

\citet{basu2021influence} empirically observed that IFs are unstable when they are applied to deep neural networks (DNNs).
The quality of influence estimation deteriorates as networks become more complex.
In this paper, we provide empirical and theoretical explanations for the instability of IFs.
We show that IFs scores are very noisy when the two data points belong to two different classes but IFs scores are much more stable when the two data points are in the same class (Sec.~\ref{sec:method}).
Based on that finding, we propose IFs-class, variants of IFs that use class information to improve the stability while introducing no additional computational cost.
IFs-class can replace IFs in anomalous data detection algorithms.
In Sec.~\ref{sec:exp}, we compare IFs-class and IFs on the error detection problem.
Experiments on various NLP tasks and datasets confirm the advantages of IFs-class over IFs.
\section{Background and Related work}
\label{sec:related}
We define the notations used in this paper.
Let $\mathbf{z} = (\mathbf{x}, \mathbf{y})$ be a data point, where $\mathbf{x} \in \mathcal{X}$ is the input, $\mathbf{y} \in \mathcal{Y}$ is the target output;
$\mathcal{Z} = \left\lbrace \mathbf{z}^{(i)} \right\rbrace_{i = 1}^n$ be a dataset of $n$ data points; 
$\mathcal{Z}_{-i} = \mathcal{Z} \backslash \mathbf{z}^{(i)}$ be the dataset $\mathcal{Z}$ with $\mathbf{z}^{(i)}$ removed;
$f_\mathbf{\bm{\theta}}: \mathcal{X} \rightarrow \mathcal{Y}$  be a model with parameter $\bm \theta$;
$\mathcal{L}_{\mathcal{Z}, \bm \theta} = \frac{1}{n} \sum_{i = 1}^n \ell(f_{{\bm{\theta}}}(\mathbf{x}^{(i)}), \mathbf{y}^{(i)}) =\frac{1}{n} \sum_{i = 1}^n \ell(\mathbf{z}^{(i)}; \bm \theta)$ be the empirical risk of $f_{\bm \theta}$ measured on $\mathcal{Z}$, where $\ell: \mathcal{Y} \times \mathcal{Y} \rightarrow \mathbb{R}^+$ is the loss function;
$\hat{\bm \theta} = \argmin_{\bm \theta} \mathcal{L}_{\mathcal{Z}, \bm \theta} $ and $\hat{\bm \theta}_{-i} = \argmin_{\bm \theta} \mathcal{L}_{\mathcal{Z}_{-i}, \bm \theta}$ be the optimal parameters of the model $f_{\bm \theta}$ trained on $\mathcal{Z}$ and $\mathcal{Z}_{-i}$.
In this paper, $f_{\bm \theta}$ is a deep network and $\hat{\bm \theta}$ is found by training $f_{\bm \theta}$ with gradient descent on the training set $\mathcal{Z}$.

\subsection{Influence function and variants}

The influence of a data point $\mathbf{z}^{(i)}$ on another data point $\mathbf{z}^{(j)}$ is defined as the change in loss at $\mathbf{z}^{(j)}$ when $\mathbf{z}^{(i)}$ is removed from the training set
\begin{align}
s^{(ij)} = \ell(\mathbf{z}^{(j)}; \hat{\bm \theta}_{-i}) - \ell(\mathbf{z}^{(j)}; \hat{\bm \theta})
\end{align}
The absolute value of $s^{(ij)}$ measures the strength of the influence of $\mathbf{z}^{(i)}$ on $\mathbf{z}^{(j)}$.
The sign of $s^{(ij)}$ show the direction of influence.
A negative $s^{(ij)}$ means that removing $\mathbf{z}^{(i)}$ decreases the loss at $\mathbf{z}^{(j)}$, i.e. $\mathbf{z}^{(i)}$ is harmful to $\mathbf{z}^{(j)}$.
$s^{(ij)}$ has high variance because it depends on a single (arbitrary) data point $\mathbf{z}^{(j)}$.
To better estimate the influence of $\mathbf{z}^{(i)}$ on the entire data distribution, researchers average the influence scores of $\mathbf{z}^{(i)}$ over a reference set $\mathcal{Z}'$
\begin{align}
s^{(i)} = \frac{1}{\abs{\mathcal{Z}'}} \sum_{\mathbf{z}^{(j)} \in \mathcal{Z}'} s^{(ij)} =\mathcal{L}_{\mathcal{Z}', \hat{\bm \theta}_{-i}} - \mathcal{L}_{\mathcal{Z}', \hat{\bm \theta}} \label{eqn:ifAverage}
\end{align}
$s^{(i)}$ is the influence of $\mathbf{z}^{(i)}$ on the reference set $\mathcal{Z}'$.
$\mathcal{Z}'$ can be a random subset of the training set or a held-out dataset.
Naive computation of $s^{(ij)}$ requires retraining $f_{\bm \theta}$ on $\mathcal{Z}_{-i}$.
\citet{koh2017understanding} proposed the influence function (IF) to quickly estimate $s^{(ij)}$ without retraining 
\begin{align}
s^{(ij)} &\approx IF(\mathbf{z}^{(i)}, \mathbf{z}^{(j)}) \nonumber \\
&\approx \frac{1}{n} \nabla_{\hat{\bm \theta}} \ell(\mathbf{z}^{(i)}; \hat{\bm \theta})^\top H_{\hat{\bm \theta}}^{-1} \nabla_{\hat{\bm \theta}} \ell(\mathbf{z}^{(j)}; \hat{\bm \theta}) \label{eqn:if}
\end{align}
where $H_{\hat{\bm\theta}} = \nicefrac{\partial^2 \mathcal{L}_{\mathcal{Z}, \hat{\bm \theta}}}{\partial {\bm \theta}^2}$ is the Hessian at $\hat{\bm\theta}$.
Exact computation of $H^{-1}_{\hat{\bm \theta}}$ is intractable for modern networks.
\citet{koh2017understanding} developed a fast algorithm for estimating  $H_{\hat{\bm \theta}}^{-1} \nabla_{\hat{\bm \theta}} \ell(\mathbf{z}^{(j)}; \hat{\bm \theta})$ and used only the derivatives w.r.t.~the last layer's parameters to improve the algorithm's speed.
\citet{charpiat2019input} proposed gradient dot product (GD) and gradient cosine similarity (GC) as faster alternatives to IF.
\citet{pruthi2020estimating} argued that the influence can be better approximated by accumulating it through out the training process (TracIn).
The formula for IFs are summarized in Tab.~\ref{tab:ifs} in Appx.~\ref{appx:formula}.

IFs can be viewed as measures of the similarity between the gradients of two data points.
Intuitively, gradients of harmful examples are dissimilar from that of normal examples (Fig.~\ref{fig:grad0}).

\subsection{Influence functions for error detection}
In the error detection problem, we have to detect data points with wrong labels.
Given a (potentially noisy) dataset $\mathcal{Z}$, we have to rank data points in $\mathcal{Z}$ by how likely they are erroneous.
Removing or correcting errors improves the performance and robustness of models trained on that dataset.

Traditional error detection algorithms that use hand designed rules \cite{chu2013holistic} or simple statistics \cite{huang2018auto}, do not scale well to deep learning datasets.
\citet{cohen2020detecting, dau2022towards} used IFs to detect adversarial and erroneous examples in deep learning datasets.
\citet{dau2022towards} used IFs to measure the influence of each data point $\mathbf{z} \in \mathcal{Z}$ on a clean reference set $\mathcal{Z}'$. 
Data points in $\mathcal{Z}$ are ranked by how harmful they are to $\mathcal{Z}'$.
Most harmful data points are re-examined by human or are removed from $\mathcal{Z}$ (Alg.~\ref{alg:simErr} in Appx.~\ref{appx:formula}).
In this paper, we focus on the error detection problem but IFs and IFs-class can be used to detect other kinds of anomalous data.

\section{Method}
\label{sec:method}

\subsection{Motivation}
\label{sec:motivation}

\begin{figure}[!ht]
\centering
\includegraphics[width=0.29\textwidth]{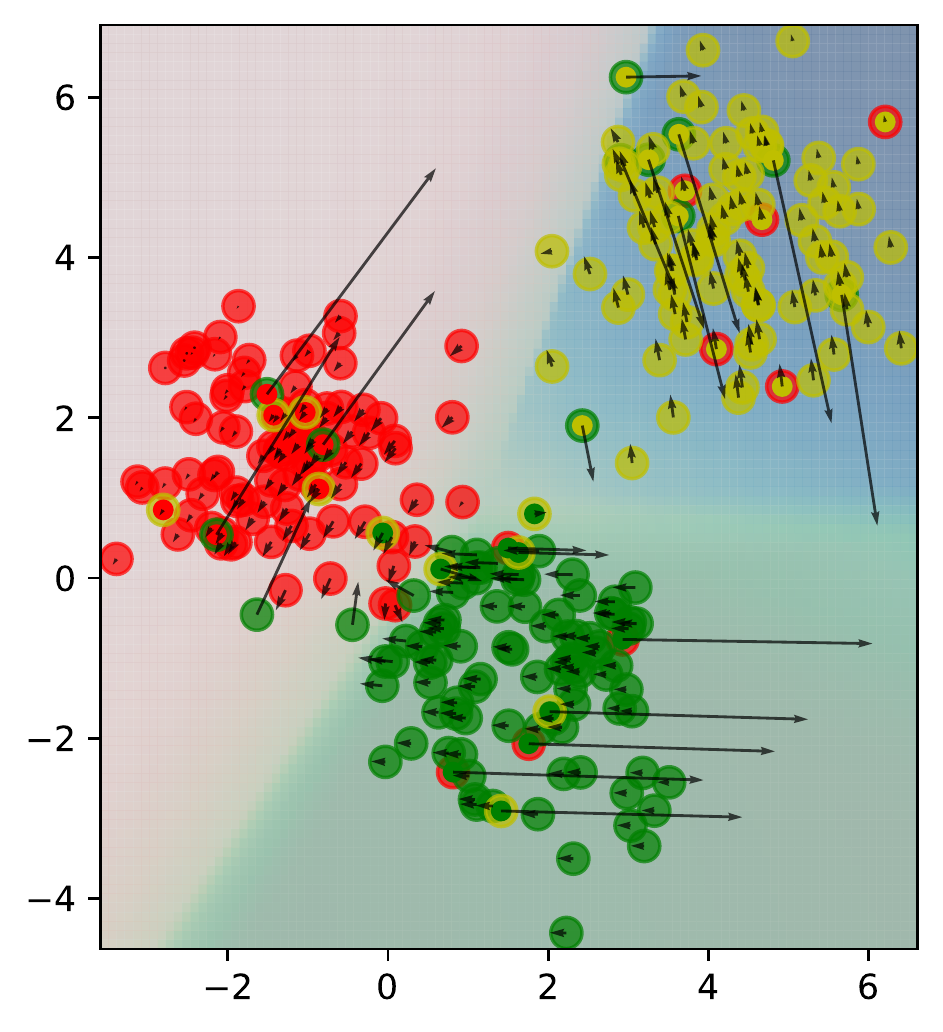}
\caption{Gradient pattern on a classification problem. 
A mislabeled data point is shown by a circle with two colors, the inner color is the original (true) class, the outer color is the new (noisy) class.
We plot only the first 2 dimensions of the gradient. 
See Appx.~\ref{appx:gradFail} for implementation details and other gradient dimensions.}
\label{fig:grad0}
\end{figure}

\begin{figure}[!hbt]
\centering
\includegraphics[width=0.32\textwidth]{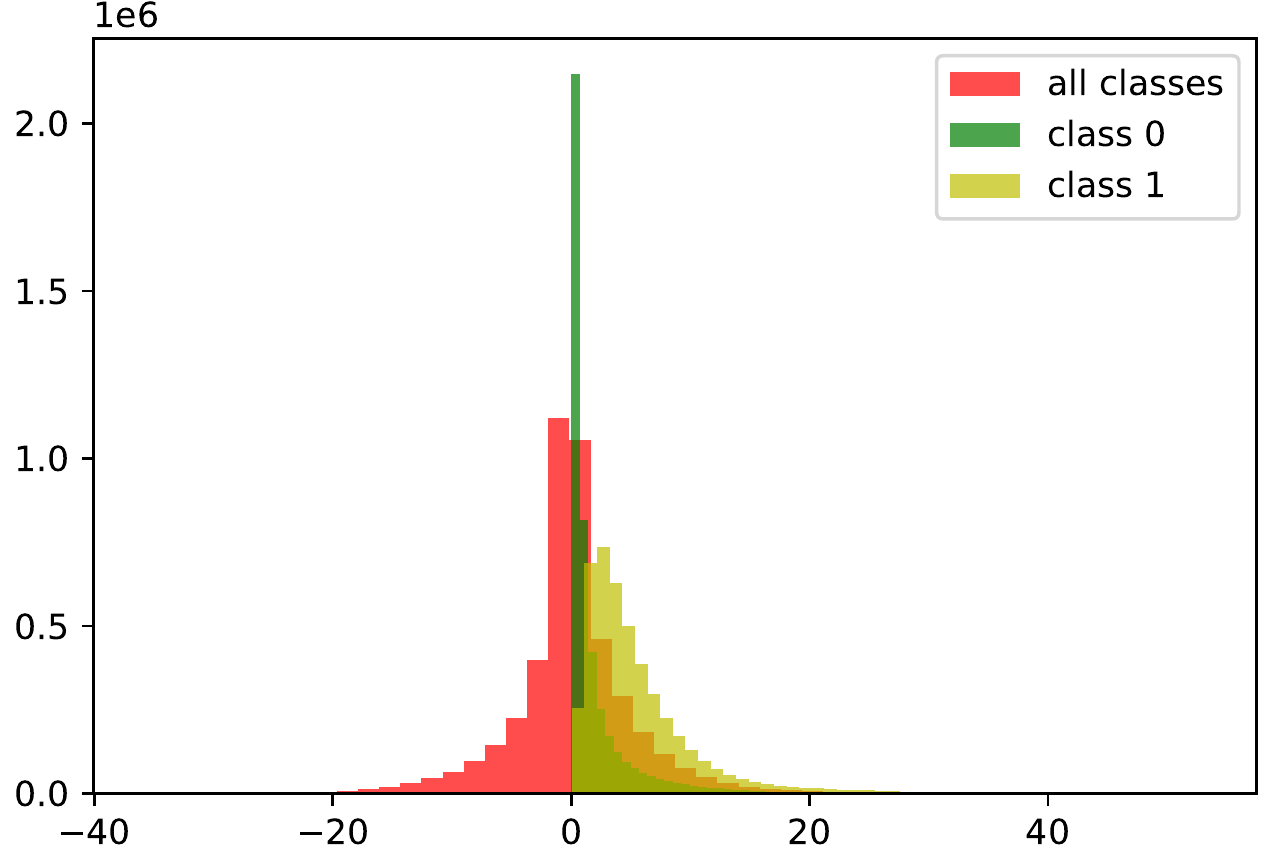}
\caption{GD score distribution on the IMDB dataset.
Results on other datasets are shown in Appx.~\ref{appx:gradFail}.
}
\label{fig:gdDist}
\end{figure}

\citet{basu2021influence} attributed the instability of IFs to the non-convexity of DNNs and the errors in Taylor's expansion and Hessian-Vector product approximation.
In this section, we show that the learning dynamics of DNNs makes examples from different classes unrelated and can have random influence on each other.

\citet{pezeshkpour-etal-2021-empirical,  hanawa2020evaluation} empirically showed that IFs with last layer gradient perform as well as or better than IFs with all layers' gradient and 
variants of IF behave similarly.
Therefore, we analyze the behavior of GD with last layer's gradient and generalize our results to other IFs.
Fig.~\ref{fig:grad0} shows the last layer's gradient of an MLP on a 3-class classification problem.
In the figure, gradients of mislabeled data points have large magnitudes and are opposite to gradients of correct data points in the true class.
However, gradients of mislabeled data points are not necessarily opposite to that of correct data points from other classes.
Furthermore, gradients of two data points from two different classes are almost perpendicular.
We make  the following observation. 
A mislabeled/correct data point often has a very negative/positive influence on data points of the same (true) class, but its influence on other classes is noisy and small.

We verify the observation on real-world datasets. (Fig.~\ref{fig:gdDist}).
We compute GD scores of pairs of clean data points from 2 different classes and plot the score's distribution.
We repeat the procedure for pairs of data points from each class.
In the 2-class case, GD scores are almost normally distributed with a very sharp peak at 0.
That means, in many cases, a clean data point from one class has no significant influence on data points from the other class.
And when it has a significant effect, the effect could be positive or negative with equal probability.
In contrast, GD scores of pairs of data points from the same class are almost always positive.
A clean data point almost certainly has a positive influence on clean data points of the same class.

Our theoretical analysis shows that when the two data points have different labels, then the sign of GD depends on two random variables, the sign of inner product of the features and the sign of inner product of gradients of the losses w.r.t.~the logits.
And as the model becomes more confident about the labels of the two data points, the magnitude of GD becomes smaller very quickly.
Small perturbations to the logits or the features can flip the sign of GD.
In contrast, if the two data points have the same label, then the sign of GD depends on only one random variable, the sign of the inner product of the feature, and the GD's magnitude remains large when the model becomes more confident.
Mathematical details are deferred to Appx.~\ref{appx:explain}.

\subsection{Class based IFs for error detection}
\label{sec:algorithm}
Our class based IFs for error detection is shown in Alg.~\ref{alg:classErr}.
In Sec.~\ref{sec:motivation}, we see that an error has a very strong negative influence on correct data points in the true class, and a correct data point has a positive influence on correct data points in the true class.
Influence score on the true class is a stronger indicator of the harmfulness of a data point and is better at differentiating erroneous and correct data points.
Because we do not know the true class of $\mathbf{z}^{(i)}$ in advance, we compute its influence score on each class in the reference set $\mathcal{Z}'$ and take the minimum of these influence scores as the indicator of the harmfulness of $\mathbf{z}^{(i)}$ (line \ref{algl:classErrFor}-\ref{algl:classErrMin}). 
Unlike the original IFs, IFs-class are not affected by the noise from other classes and thus, have lower variances (Fig.~\ref{fig:scoreHistDistance} in Appx.~\ref{appx:formula}).
In Appx.~\ref{appx:formula}, we show that our algorithm has the same computational complexity as IFs based error detection algorithm.

\begin{algorithm}[t!]
\caption{Class based influence function for error detection.}\label{alg:classErr}
\begin{algorithmic}[1]
\Require \\
$\mathcal{Z} = \left\lbrace \mathbf{z}^{(i)} \right\rbrace_{i=1}^n$: a big noisy dataset\\ 
$C$: number of classes \\
$\mathcal{Z}'_k = \left\lbrace \mathbf{z}'^{(j_k)} \right\rbrace_{j_k=1}^{m_k}$: clean data from class $k$ \\
$\mathcal{Z}' = \bigcup_{k=1}^C \mathcal{Z}'_k$: a clean reference dataset \\
$f_{\hat{\bm{\theta}}}$: a deep model pretrained on $\mathcal{Z}$ \\
$\textnormal{sim}(\cdot, \cdot)$: a similarity measure in Tab.~\ref{tab:ifs}
\Ensure $\hat{\mathcal{Z}}$: data points in $\mathcal{Z}$ ranked by score

\For{$\mathbf{z}^{(i)} \in \mathcal{Z}$}
\For{$k = 1, ..., C$} \label{algl:classErrFor}
	\State ${s^{(i)}_k=\frac{1}{m_k} \sum_{j = 1}^{m_k} \textnormal{sim}(\nabla_{\hat{\bm{\theta}}}\ell(\mathbf{z}^{(i)}), \nabla_{\hat{\bm{\theta}}}\ell(\mathbf{z}'^{(j_k)}))}$
\EndFor \label{algl:classErrEndFor}
\State $s^{(i)} = \min_k(s^{(i)}_k)$ \label{algl:classErrMin}
\EndFor\\
$\hat{\mathcal{Z}} = \textnormal{sort}(\mathcal{Z}, \textnormal{key}=\bm s, \textnormal{ascending}=\textnormal{True})$ \\
\Return $\hat{\mathcal{Z}}$

\end{algorithmic}
\end{algorithm}

\section{Experiments}
\label{sec:exp}

%

\begin{figure*}
\centering
\subfigure[]{\includegraphics[width=0.3\textwidth]{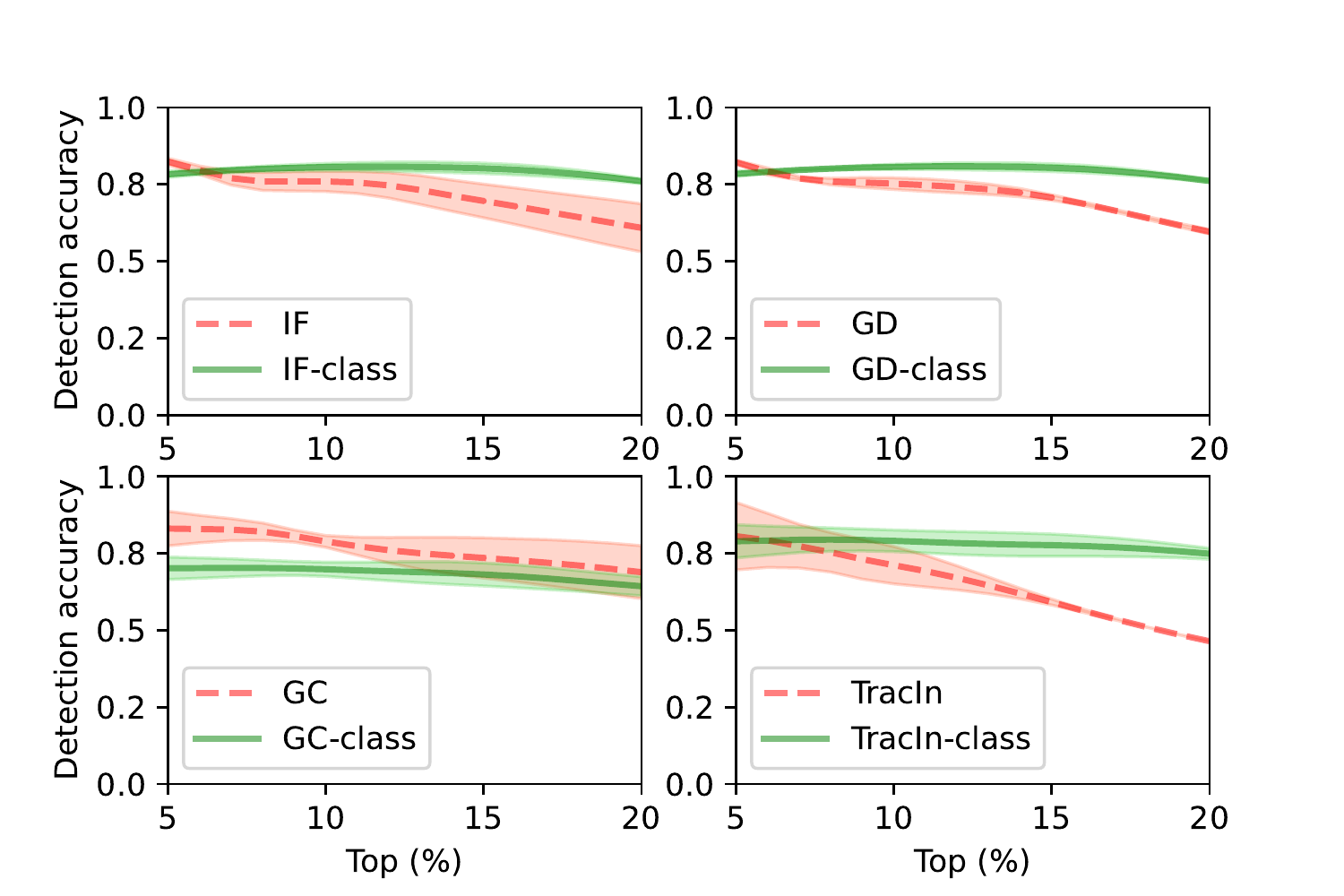} \label{fig:snli}
}
\subfigure[]{\includegraphics[width=0.3\textwidth]{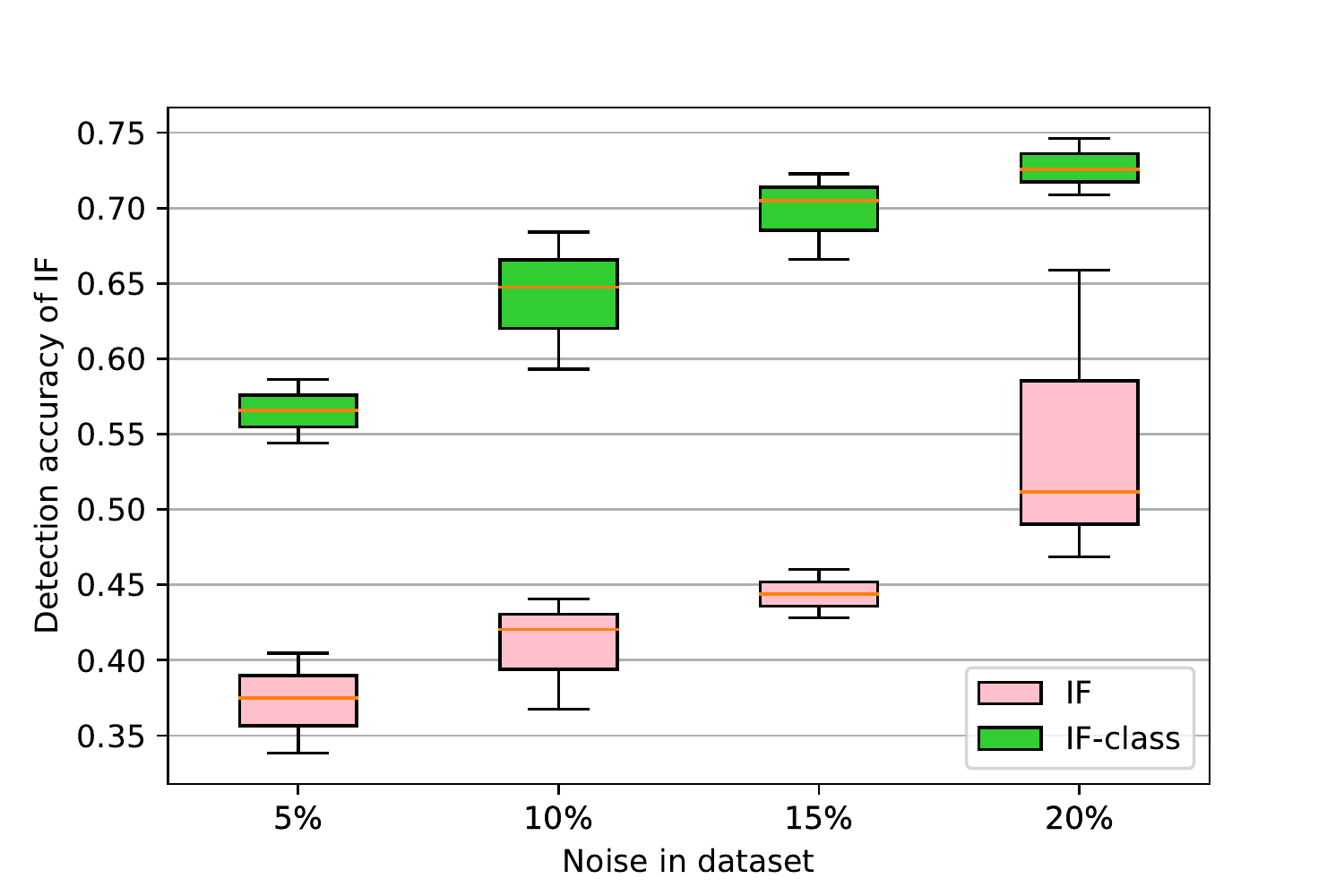} \label{fig:imdbChange}
}
\subfigure[]{\includegraphics[width=0.27\textwidth]{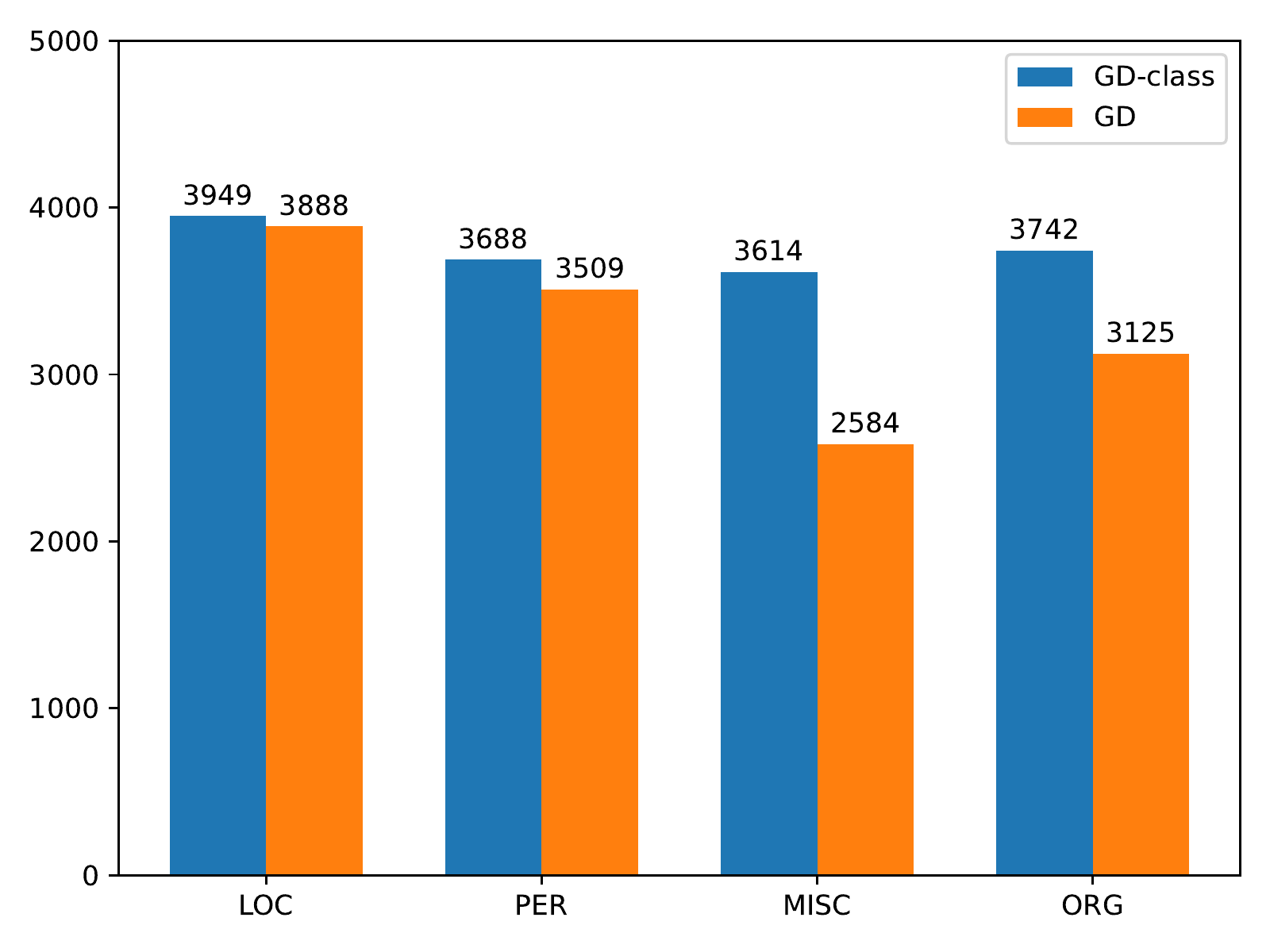} \label{fig:nerCats}
}
\caption{\subref{fig:snli} Error detection accuracy on SNLI dataset with $p=20\%$.
\subref{fig:imdbChange} Error detection accuracy of IF and IF-class on IMDB dataset with different values of $p$.
\subref{fig:nerCats} Number of erroneous NER tokens detected by GD and GD-class at $p=30\%, r=30\%, q=9\%$, grouped by entity types of the erroneous tokens.}
\end{figure*}


\noindent \textbf{Experiment setup}
We evaluate the error detection performance of IFs-class on 2 NLP tasks, (1) text classification on IMDB \cite{imdb}, SNLI \cite{snli}, and BigCloneBench \cite{bigclone} datasets, and (2) NER on the CoNLL2003 \cite{tjong-kim-sang-de-meulder-2003-introduction} dataset.
For text classification tasks, we detect text segments with wrong labels.
For the NER task, we detect tokens with wrong entity types.
We use BERT \cite{devlin2018bert} and CodeBERT \cite{feng-etal-2020-codebert} in our experiments.
Implementation details are located in Appx.~\ref{appx:data}.
To create benchmark datasets $\mathcal{Z}$'s, we inject random noise into the above datasets. 
For text classification datasets, we randomly select $p\%$ of the data points and change their labels to other random classes.
For the CoNLL-NER dataset, we randomly select $p\%$ of the sentences and change the labels of $r\%$ of the phrases in the selected sentences.
All tokens in a selected phrase are changed to the same class.
The reference set $\mathcal{Z}'$ is created by randomly selecting $m_k$ clean data points from each class in $\mathcal{Z}$.
Models are trained on the noisy dataset $\mathcal{Z}$.
To evaluate an error detection algorithm, we select top $q\%$ most harmful data points from the sorted dataset $\hat{\mathcal{Z}}$ and check how many percent of the selected data points are really erroneous.
Intuitively, increasing $q$ allows the algorithm to find more errors (increase recall) but may decrease the detection accuracy (decrease precision).

\noindent \textbf{Result and Analysis} Because results on all datasets share the same patterns, we report representative results here and defer the full results to Appx.~\ref{appx:gradFail}. 

Fig.~\ref{fig:snli} shows the error detection accuracy on the SNLI dataset and how the accuracy changes with $q$.
Except for the GC algorithm, our class-based algorithms have higher accuracy and lower variance than the non-class-based versions.
When $q$ increases, the performance of IFs-class does not decrease as much as that of IFs.
This confirms that IFs-class are less noisy than IFs.
Class information fails to improve the performance of GC.
To understand this, let's reconsider the similarity measure $\textnormal{sim}(\cdot, \cdot)$.
Let's assume that there exist some clean data points $\mathbf{z}'^{(j)} \in \mathcal{Z}'$ with a very large gradient $\nabla_{\hat{\bm \theta}}\ell(\mathbf{z}'^{(j)})$.
If the similarity measure does not normalize the norm of $\nabla_{\hat{\bm \theta}}\ell(\mathbf{z}'^{(j)})$, then $\mathbf{z}'^{(j)}$ will have the dominant effect on the influence score.
The noise in the influence score is mostly caused by these data points.
GC normalizes both gradients, $\nabla_{\hat{\bm \theta}}\ell(\mathbf{z}^{(i)})$ and $\nabla_{\hat{\bm \theta}}\ell(\mathbf{z}'^{(j)})$, and effectively removes such noise.
However, gradients of errors tend to be larger than that of normal data points (Fig.~\ref{fig:grad0}).
By normalizing both gradients, GC removes the valuable information about magnitudes of gradients of errors $\nabla_{\hat{\bm{\theta}}}\ell(\mathbf{z}^{(i)})$.
That lowers the detection performance.
In Fig.~\ref{fig:snli}, we see that the performance of GC when $q \ge 15\%$ is lower than that of other class-based algorithms.
Similar trends are observed on other datasets (Fig.~\ref{fig:appxImdb}, \ref{fig:appxBigClone}, \ref{fig:appxCoNLL} in Appx.~\ref{appx:gradFail}).

Fig.~\ref{fig:imdbChange} shows the change in detection accuracy as the level of noise $p$ goes from $5\%$ to $20\%$.
For each value of $p$, we set $q$ to be equal to $p$.
Our class-based influence score significantly improves the performance and reduces the variance.
We note that when $p$ increases, the error detection problem becomes easier as there are more errors.
The detection accuracy, therefore, tends to increase with $p$ as shown in Fig.~\ref{fig:imdbChange}, \ref{fig:bigCloneChange}, \ref{fig:snliChange}.

Fig.~\ref{fig:nerCats} shows that GD-class outperforms GD on all entity types in CoNLL2003-NER.
The performance difference between GD-class and GD is greater on the MISC and ORG categories.
Intuitively, a person's name can likely be an organization's name but the reverse is less likely.
Therefore, it is harder to detect that a PER or LOC tag has been changed to ORG or MISC tag than the reverse.
The result shows that IFs-class is more effective than IFs in detecting hard erroneous examples.
\section{Conclusion}
\label{sec:conclusion}

In this paper, we study influence functions and identify the source of their instability.
We give a theoretical explanation for our observations.
We introduce a stable variant of IFs and use  that to develop a high performance error detection algorithm.
Our findings shed light of the development of new influence estimators and on the application of IFs in downstream tasks.

\clearpage
\pagebreak
\section*{Limitations}
Our paper has the following limitations
\begin{enumerate}
\item Our class-based influence score cannot improve the performance of GC algorithm.
Although class-based version of GD, IF, and TracIn outperformed the original GC, we aim to develop a stronger version of GC.
From the analysis in Sec.~\ref{sec:exp}, we believe that a partially normalized GC could have better performance.
In partial GC, we normalize the gradient of the clean data point $\mathbf{z}'^{(j)}$ only.
That will remove the noise introduced by $\lVert{\nabla_{\hat{\bm \theta}}\ell(\mathbf{z}'^{(j)})}\rVert$ while retaining the valuable information about the norm of $\nabla_{\hat{\bm \theta}}\ell(\mathbf{z}^{(i)})$.
\end{enumerate}
\section*{Ethics Statement}
Our paper consider a theoretical aspect of influence functions.
It does not have any biases toward any groups of people.
Our findings do not cause any harms to any groups of people.


\clearpage
\pagebreak
\bibliography{references}
\bibliographystyle{acl_natbib}

\appendix

\section{Additional algorithms and formula}
\label{appx:formula}

\begin{table}[!h]
\caption{Influence function and its variants. We drop the constant factor $\nicefrac{1}{n}$ for clarity.}
\centering
\begin{tabular}{l|l}
IF & $\nabla_{\hat{\bm \theta}} \ell(\mathbf{z}^{(i)}; \hat{\bm \theta})^\top H_{\hat{\bm \theta}}^{-1} \nabla_{\hat{\bm \theta}} \ell(\mathbf{z}^{(j)}; \hat{\bm \theta})$ \\
GD & $\left \langle \nabla_{\hat{\bm \theta}} \ell(\mathbf{z}^{(i)}), \nabla_{\hat{\bm \theta}} \ell(\mathbf{z}^{(j)}) \right \rangle$ \\
GC & $\cos( \nabla_{\hat{\bm \theta}} \ell(\mathbf{z}^{(i)}), \nabla_{\hat{\bm \theta}} \ell(\mathbf{z}^{(j)}))
$ \\
TracIn & $ \sum_{t=1}^T \eta_t \left\langle \nabla_{\bm \theta^{(t)}} \ell(\mathbf{z}^{(i)}), \nabla_{\bm \theta^{(t)}} \ell(\mathbf{z}^{(j)}) \right\rangle$\\
\end{tabular}
\label{tab:ifs}
\end{table}

\begin{algorithm}[hbt!]
\caption{Influence function based error detection \cite{dau2022towards}} \label{alg:simErr}
\begin{algorithmic}[1]
\Require \\
$\mathcal{Z} = \left\lbrace \mathbf{z}^{(i)} \right\rbrace_{i=1}^n$: a big noisy dataset\\ 
$\mathcal{Z}' = \left\lbrace \mathbf{z}'^{(j)} \right\rbrace_{j=1}^m$: a small reference dataset \\
$f_{\hat{\bm{\theta}}}$: a deep model pretrained on $\mathcal{Z}$ \\
$\textnormal{sim}(\cdot, \cdot)$: a similarity measure in Tab.~\ref{tab:ifs}
\Ensure $\hat{\mathcal{Z}}$: data points in $\mathcal{Z}$ ranked by score

\For{$\mathbf{z}^{(i)} \in \mathcal{Z}$}
\State $s^{(i)} = \frac{1}{m} \sum_{j = 1}^m \textnormal{sim}(\nabla_{\hat{\bm{\theta}}}\ell(\mathbf{z}^{(i)}), \nabla_{\hat{\bm{\theta}}}\ell(\mathbf{z}'^{(j)}))$\label{alg:if}
\EndFor\\
$\hat{\mathcal{Z}} = \textnormal{sort}(\mathcal{Z}, \textnormal{key}=\bm s, \textnormal{ascending}=\textnormal{True})$ \\
\Return $\hat{\mathcal{Z}}$

\end{algorithmic}
\end{algorithm}

\subsection*{Computational complexity of error detection algorithms}
The inner for-loop in Alg.~\ref{alg:classErr} calculates $C$ influence scores.
It calls to the scoring function $\textnormal{sim}()$ exactly $\vert \mathcal{Z}' \vert = m$ times.
The complexity of the inner for-loop in Alg.~\ref{alg:classErr} is equal to that of line \ref{alg:if} in Alg.~\ref{alg:simErr}.
Thus, the complexity of Alg.~\ref{alg:classErr} is equal to that of Alg.~\ref{alg:simErr}.

\begin{figure}[!hbt]
\includegraphics[width=0.45\textwidth]{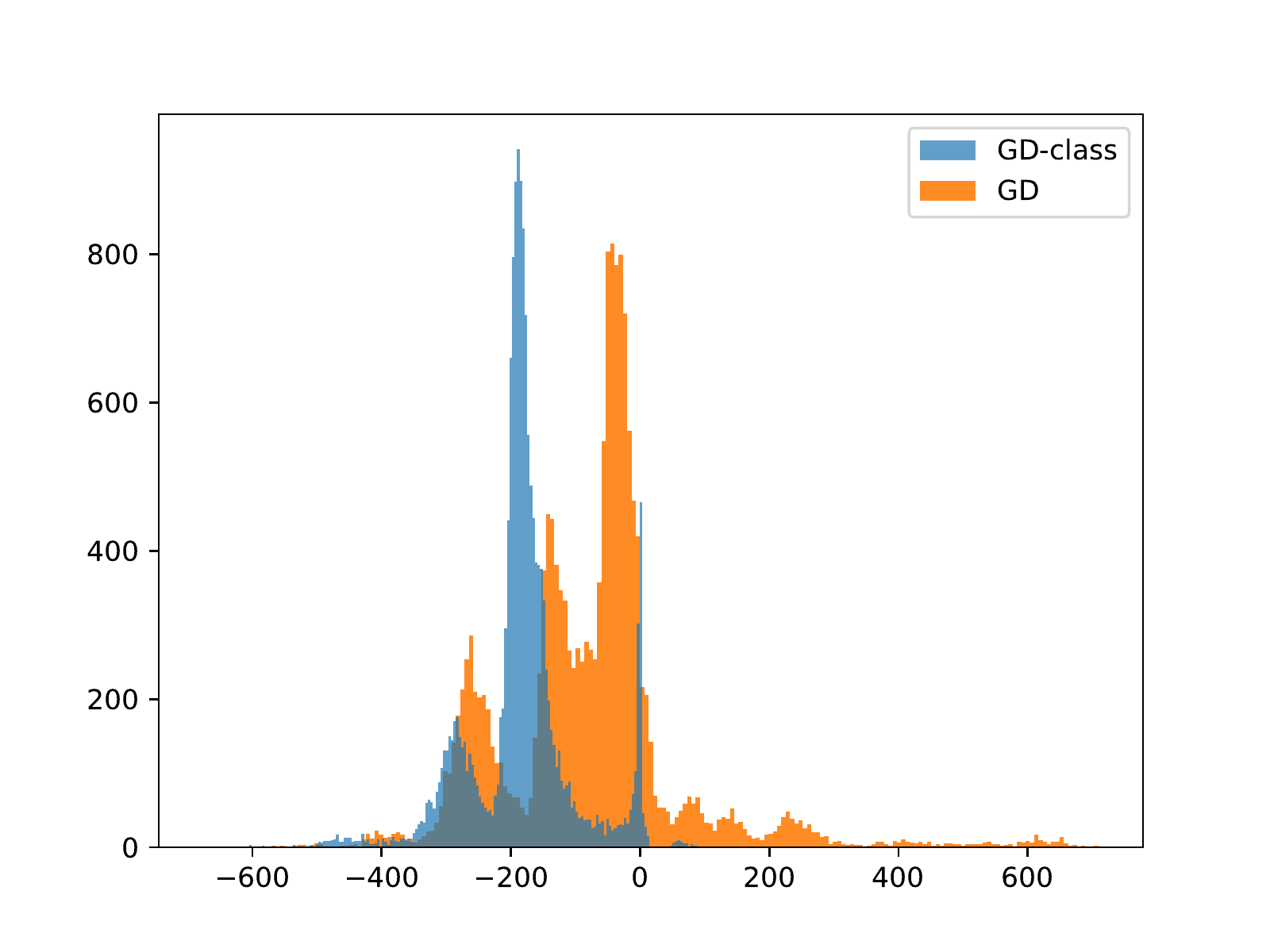}
\caption{Distributions of GD and GD-class scores of erroneous tokens in the CoNLL2003 dataset. 
GD-class scores are more concentrated and have mostly negative values.
GD scores are more spread out and the values are less negative. 
Furthermore, a significant portion of GD scores are greater than 0, i.e. GD `thinks' that these erroneous data points have positive influence on clean data points in $\mathcal{Z}'$.
In contrast, GD-class scores are more concentrated and almost always have negative values.
This shows a clear advantage of GD-class over GD.}
\label{fig:scoreHistDistance}
\end{figure}

\section{Implementation details}
\label{appx:data}

\subsection{Experiment setup}
We used standard datasets and models and experimented with 5 different random seeds and reported the mean and standard deviation.
A Nvidia RTX 3090 was used to run our experiments. 
Models are trained with the AdamW optimizer \cite{loshchilov2017decoupled} with learning rate $\eta = 5e-5$,
cross entropy loss function, and batch-size of 16.
The epoch with the best classification accuracy on the validation set was used for error detection.

Our source code and guidelines were attached to the supplementary materials.

\subsection{Datasets}
\label{datasets}
\noindent \textbf{IMDB} \cite{imdb} The dataset includes 50000 reviews from the Internet Movie Database (IMDb) website. 
The task is a binary sentiment analysis task.
The dataset contains an even number of positive and negative reviews. 
The IMDB dataset is split into training, validation, and test sets of sizes 17500, 7500, and 25000.
The IMDB dataset can be found at \url{https://ai.stanford.edu/~amaas/data/sentiment/}

\noindent \textbf{SNLI} dataset (Standart Natural Language Inference) \cite{snli} consists of 570k sentence pairs manually labeled as entailment, contradiction, and neutral. We convert these labels into numbers. It is geared towards serving as a benchmark for evaluating text representational systems.
This dataset is available at \url{https://nlp.stanford.edu/projects/snli/}.

\noindent \textbf{BigCloneBench} \cite{bigclone} is a huge code clone benchmark that includes over 6,000,000 true clone pairs and 260,000 false clone pairs from 10 different functionality. The task is to predict whether two pieces of code have the same semantics. 
This dataset is commonly used in language models for code \cite{feng-etal-2020-codebert, lu2021codexglue, guo2020graphcodebert}.
This dataset is available at \url{https://github.com/clonebench/BigCloneBench}

\noindent \textbf{CoNLL2003} \cite{tjong-kim-sang-de-meulder-2003-introduction} is one of the most influential corpora for NER model research. A large number of publications, including many landmark works, have used this corpus as a source of ground truth for NER tasks. The data consists two languages: English and German. In this paper, we use CoNLL2003 English dataset. 
The sizes of training, validation, and test are 14,987,  3,466, and 3,684 sentences correspond to 203,621, 51,362, and 46,435 tokens, respectively.
The dataset is available at \url{https://www.clips.uantwerpen.be/conll2003/ner/}

\subsection{Models}
\label{models}
\textbf{BERT} \cite{devlin2018bert} stands for Bidirectional Encoder Representations from Transformers, is based on Transformers. The BERT model in this paper was pre-trained for natural language processing tasks. We use BERT for IMDB and SNLI datasets. At the same time, we also use the BERT model for the NER problem on the CoNLL2003 dataset.

\noindent \textbf{CodeBERT} \cite{feng-etal-2020-codebert} is a bimodal pre-trained model for programming and natural languages. We use CodeBERT for BigCloneBench dataset. 
\section{Additional results}
\label{appx:gradFail}
\subsection{3-class classification experiment}
\begin{figure}[!ht]
\centering
\includegraphics[width=0.4\textwidth]{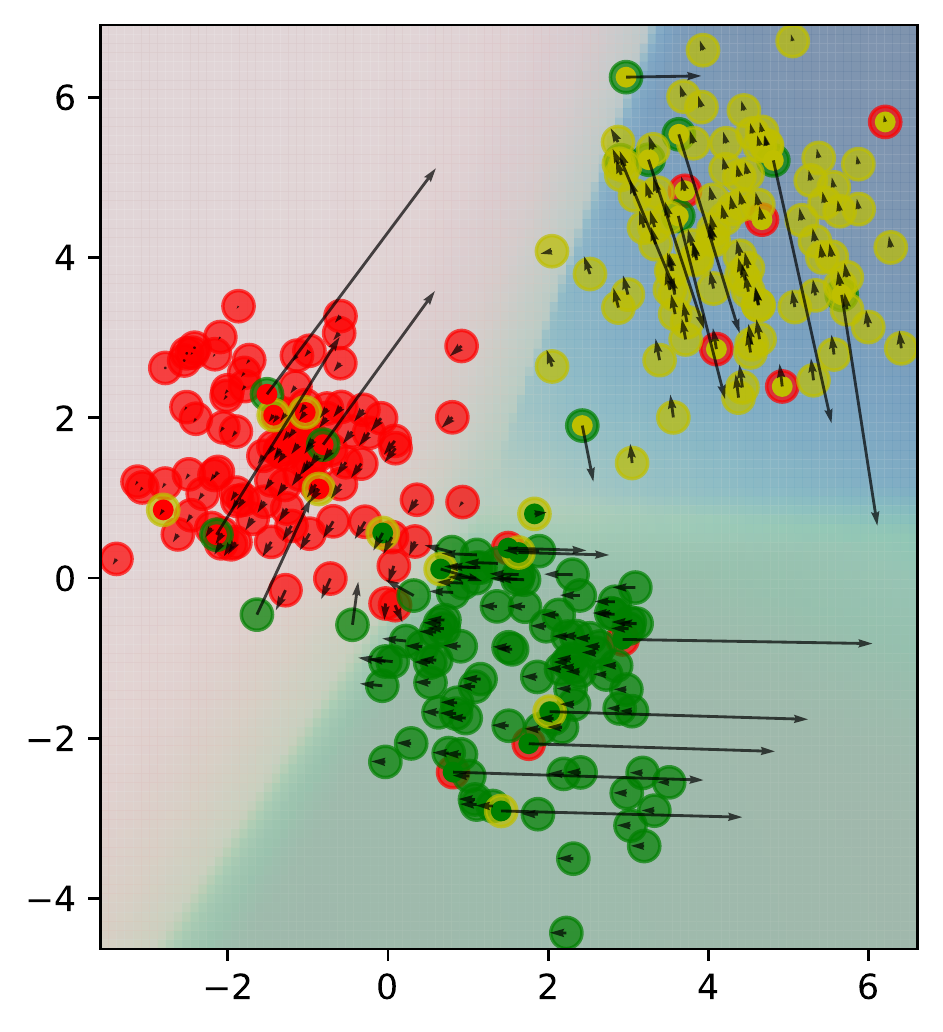}
\includegraphics[width=0.4\textwidth]{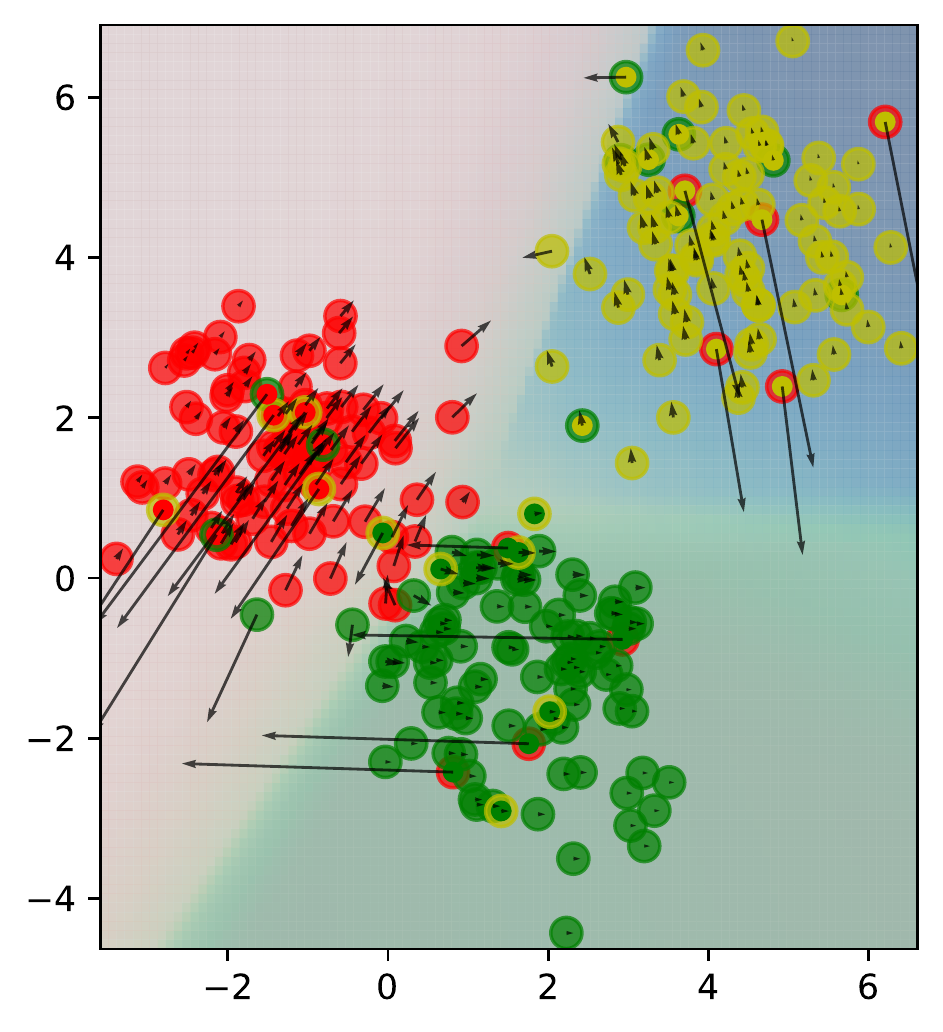}
\includegraphics[width=0.4\textwidth]{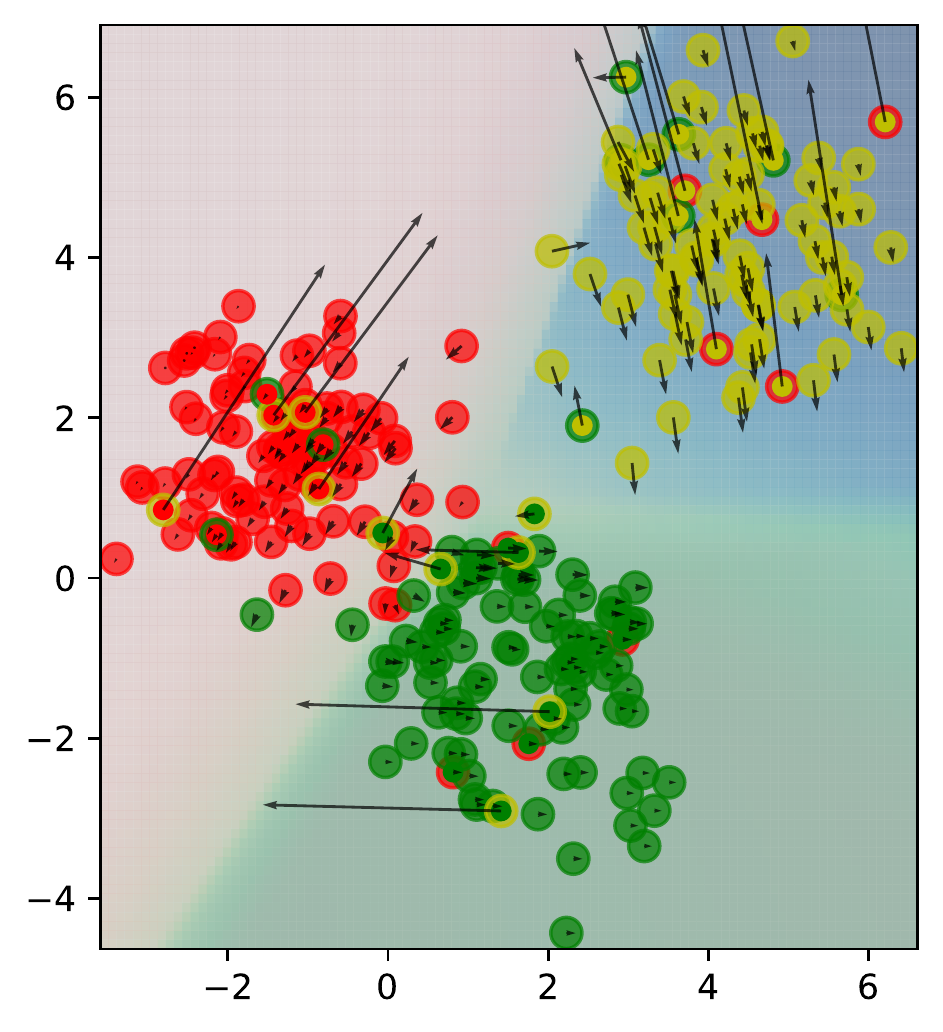}
\caption{Gradient pattern on a classification problem. Each subfigure shows 2 dimensions of the gradient. The top subfigure shows the 1st and 2nd dimensions of the gradient. The middle subfigure shows the 3rd and 4th dimensions of the gradient. The bottom subfigure shows the 5th and 6th dimensions of the gradient.}
\label{fig:appxGradFail}
\end{figure}

We train a MLP with 2 input neurons, 100 hidden neurons in the first hidden layer, 2 hidden neurons in the second hidden layer, and 3 output neurons with SGD for 1000 epochs.
The activation function is LeakyReLU and the learning rate is $\eta = 1e-3$.
The last layer has 6 parameters organized into a $3 \times 2$ matrix.
The gradient of the loss with respect to the last layer's parameters is also organized into a $3 \times 2$ matrix.
We visualize 3 rows of the gradient matrix in 3 subfigures (Fig.~\ref{fig:appxGradFail}).

\subsection{Result on IMDB, SNLI, BigCloneBench, and CoNLL2003}
To ensure a fair comparison between our class-based algorithm and algorithm \ref{alg:simErr}, we use the same reference dataset $\mathcal{Z}'$ for both algorithms.
The reference dataset $\mathcal{Z}'$ consists of $C$ classes.
We have $C = 2$ for the IMDB dataset, $C = 3$ for the SNLI dataset, $C = 2$ for the BigCloneBench dataset, and $C = 5$ for the CoNLL2003-NER dataset.
From each of the $C$ classes, we randomly select $m_k = 50$  $k = 1, ..., C$ clean data points to form $\mathcal{Z}'$.
We tried varying $m_k$ from 10 to 1000 and observed no significant changes in performance.

\begin{figure}[!h]
\centering
\includegraphics[width=0.45\textwidth]{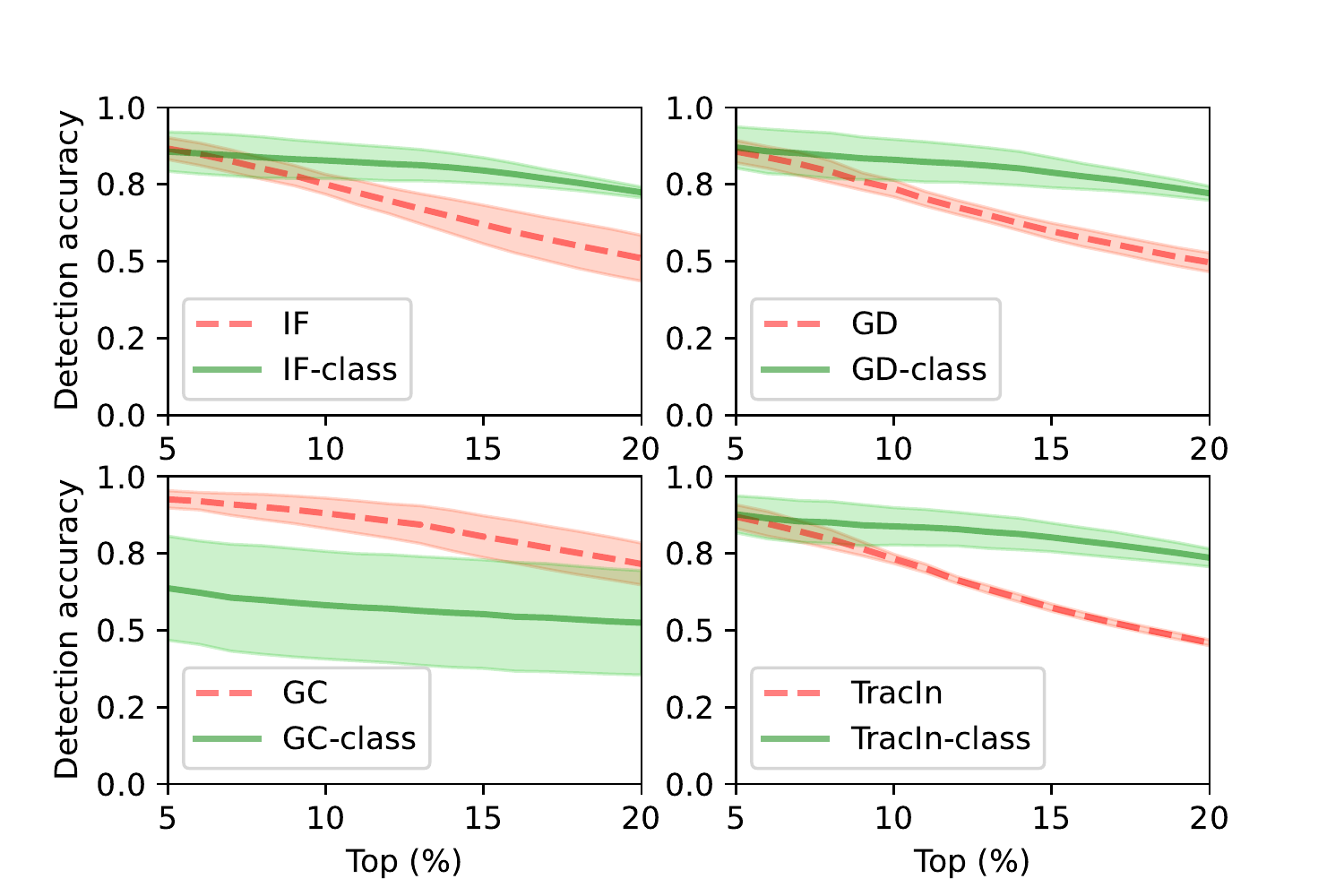}
\caption{Error detection accuracy on IMDB dataset with $p=20\%$.}
\label{fig:appxImdb}
\end{figure}

\begin{figure}[!h]
\centering
\includegraphics[width=0.45\textwidth]{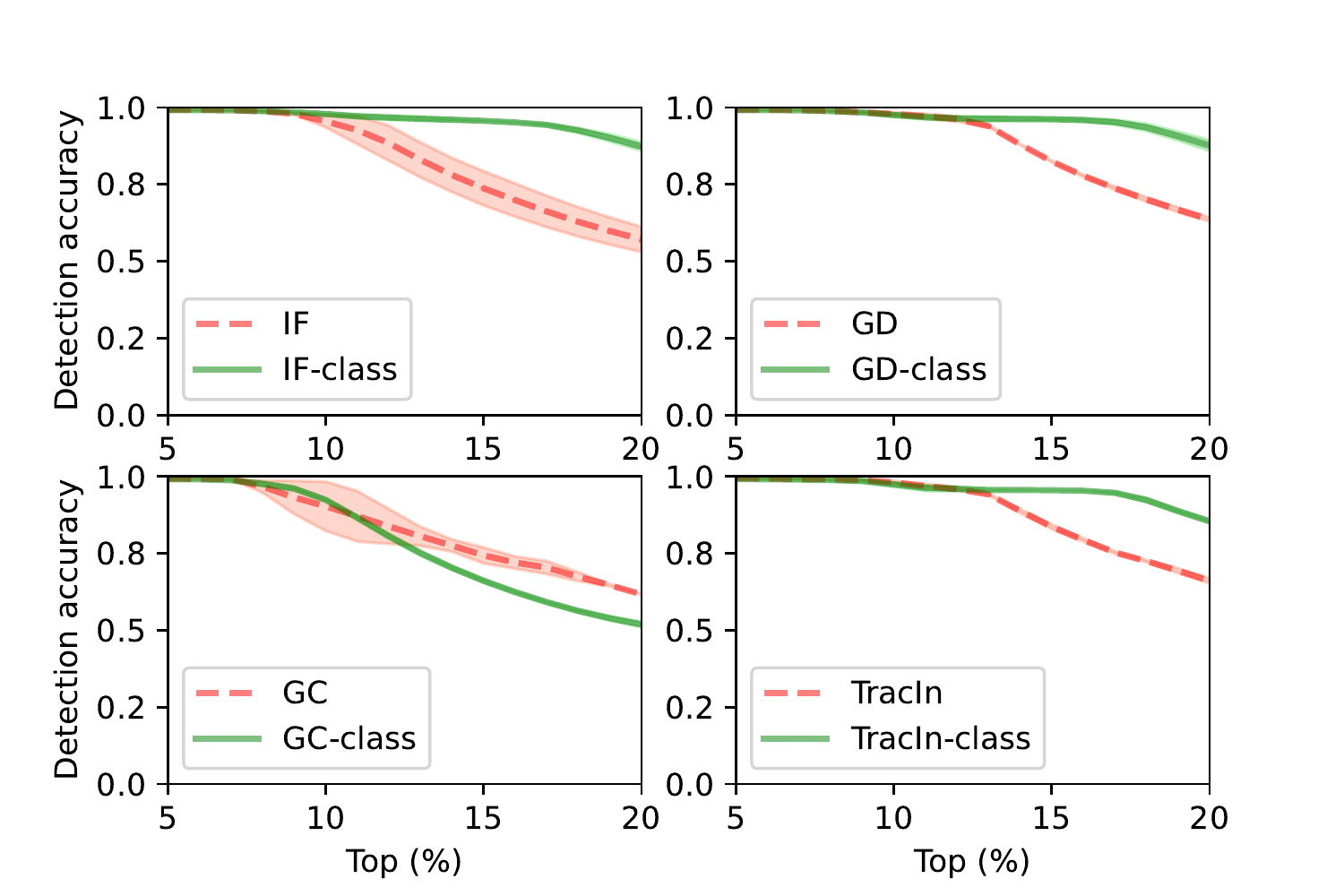}
\caption{Error detection accuracy on BigCloneBench dataset with $p=20\%$.}
\label{fig:appxBigClone}
\end{figure}

\begin{figure}[!h]
\centering
\includegraphics[width=0.45\textwidth]{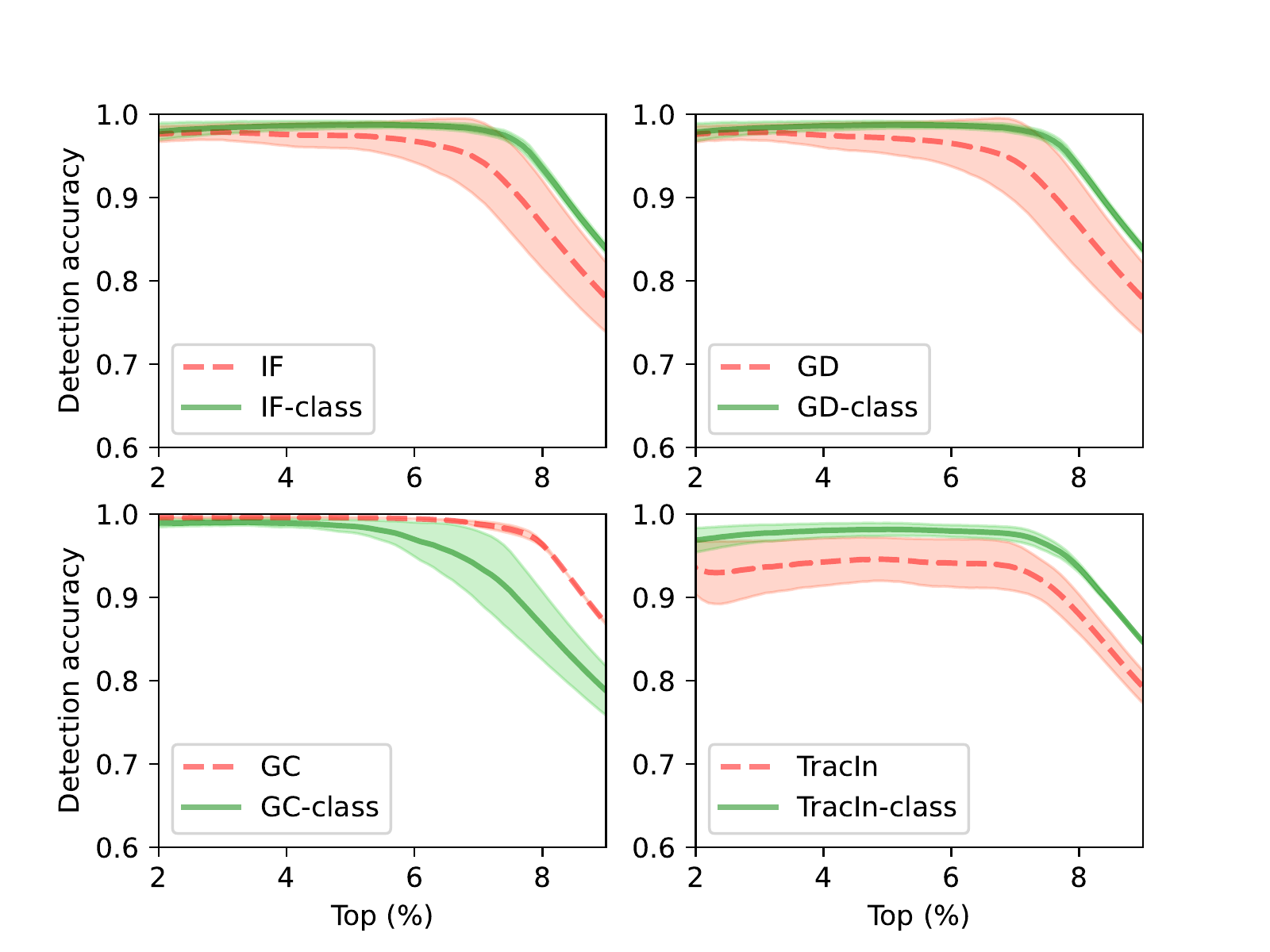}
\caption{Error detection accuracy on CoNLL2003 dataset with $p=30\%$ and $r=30\%$}
\label{fig:appxCoNLL}
\end{figure}

\begin{figure}[!h]
\centering
\includegraphics[width=0.45\textwidth]{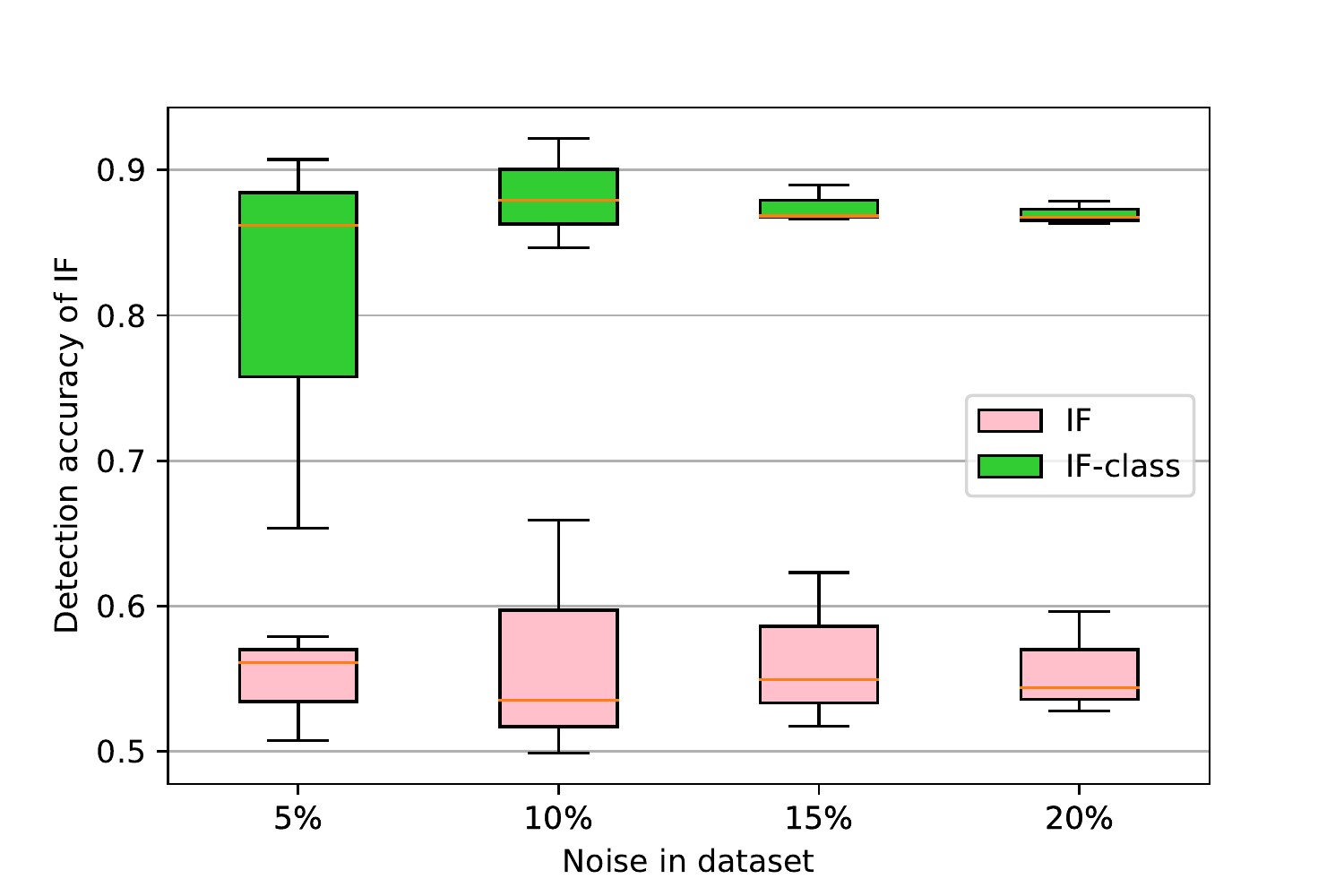}
\caption{Change in error detection accuracy on the BigCloneBench dataset as the level of noise changes.}
\label{fig:bigCloneChange}
\end{figure}

\begin{figure}[!h]
\centering
\includegraphics[width=0.45\textwidth]{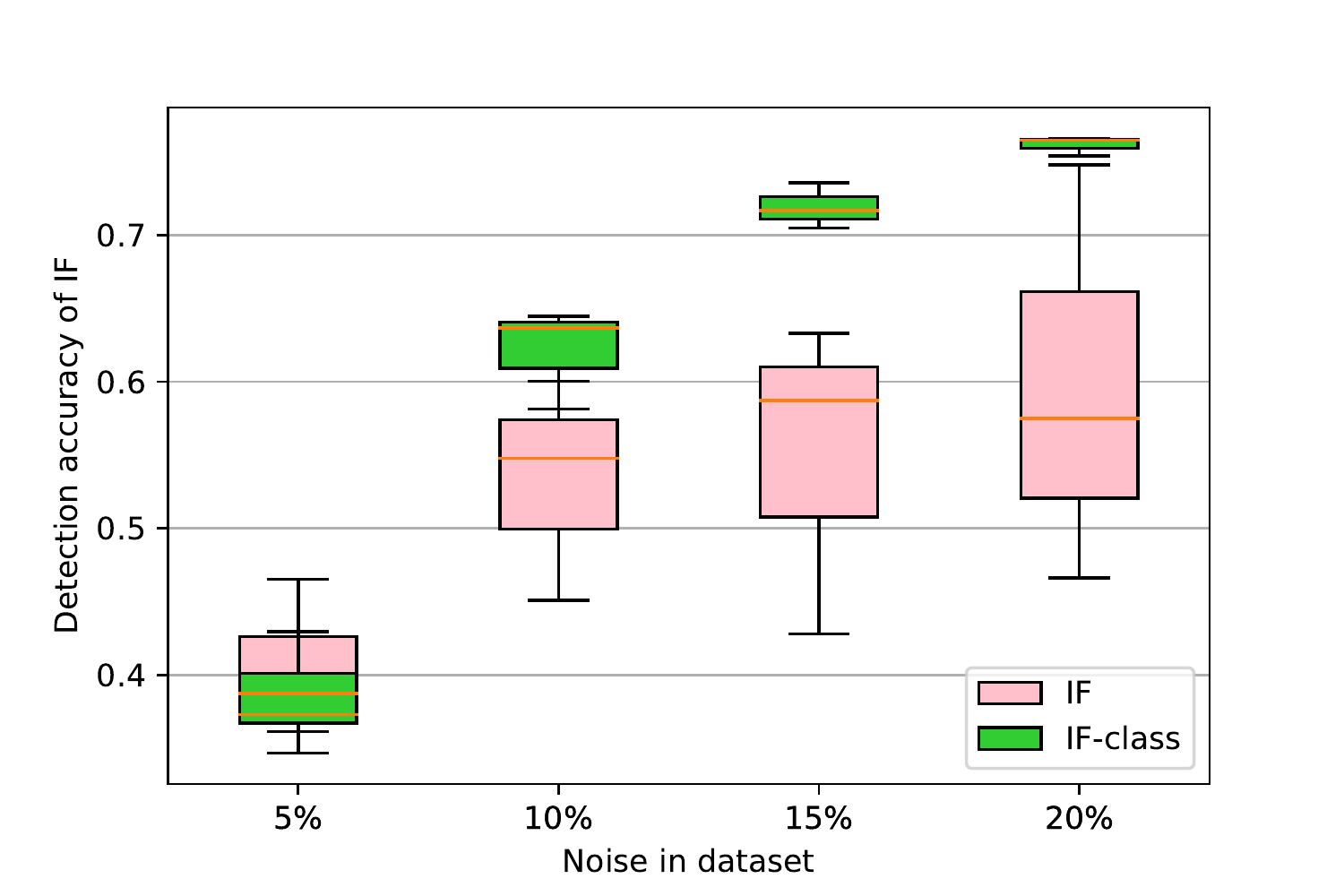}
\caption{Change in error detection accuracy on the SNLI dataset as the level of noise changes.}
\label{fig:snliChange}
\end{figure}

\begin{figure}[!ht]
\centering
\includegraphics[width=0.4\textwidth]{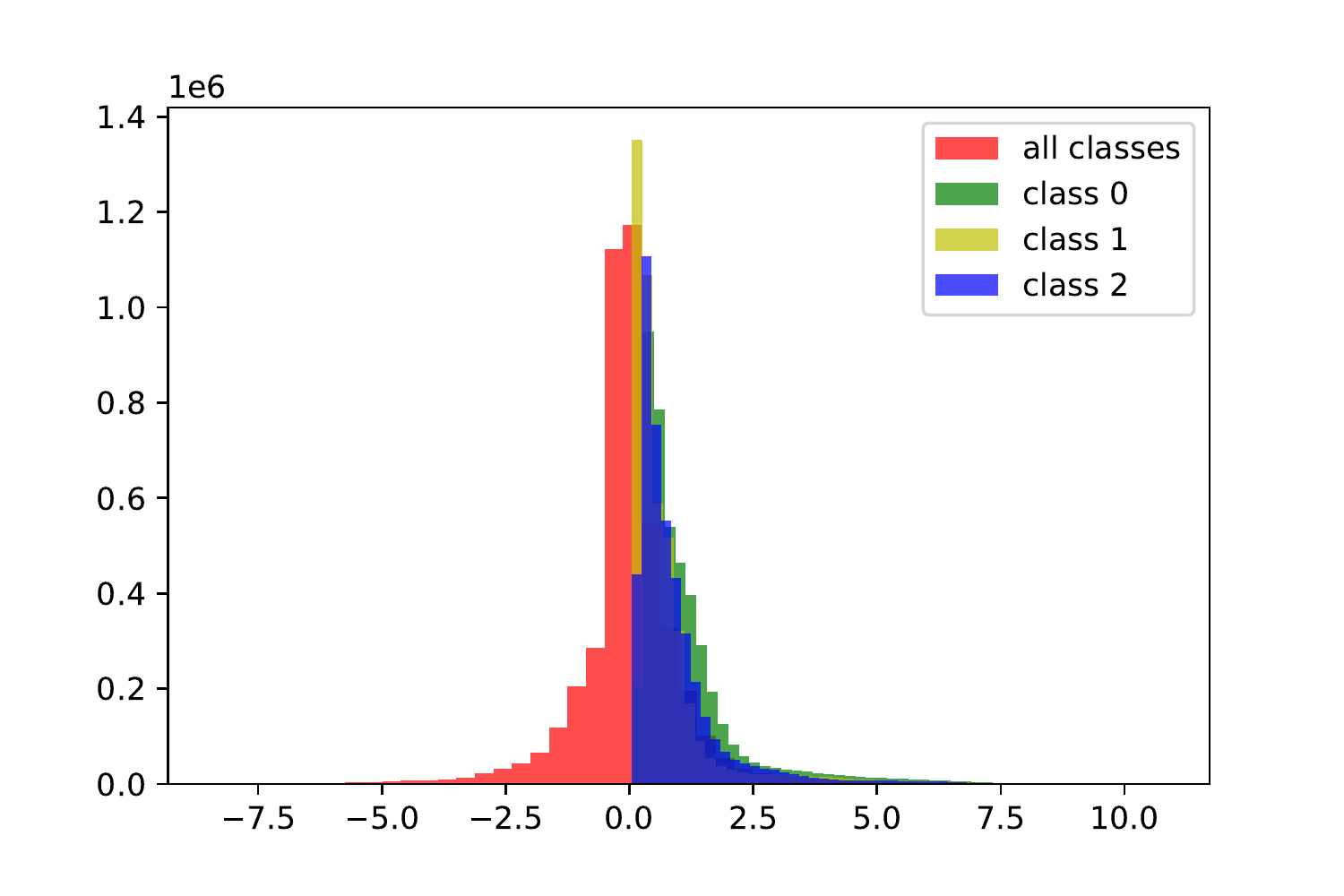}
\caption{GD score distribution on the SNLI dataset.
}
\end{figure}

\begin{figure}[!ht]
\centering
\includegraphics[width=0.4\textwidth]{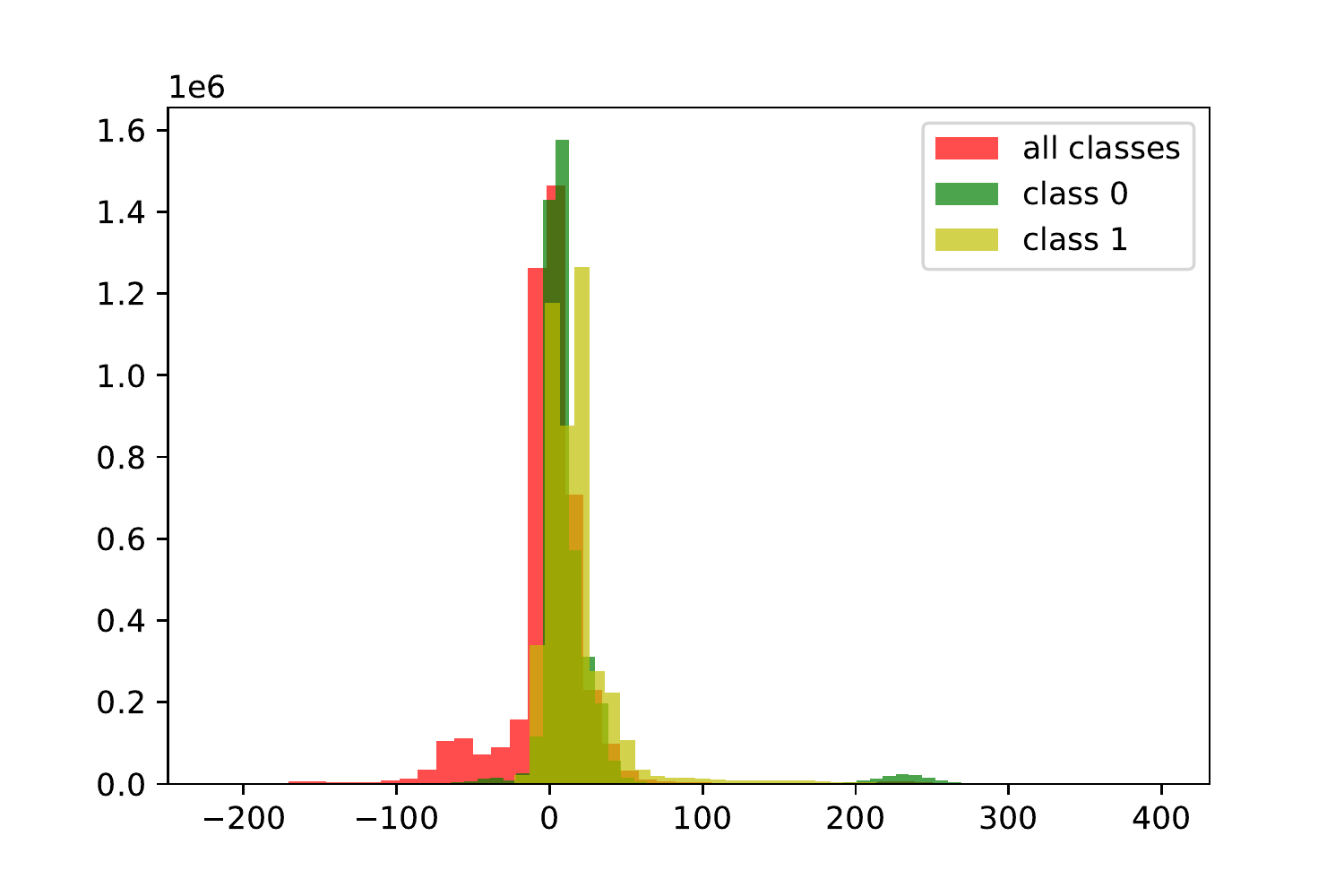}
\caption{GD score distribution on the BigCloneBench dataset.
}
\end{figure}

\clearpage

\section{Explanation of the observation in Sec.~\ref{sec:method}}
\label{appx:explain}

Let's consider a classification problem with cross entropy loss function 
\begin{align*}
\ell(\hat{\mathbf{y}}, \mathbf{y}) = \sum_{i = 1}^{d_y} y_i \log \hat{y}_i
\end{align*}
where $d_y$ is the number of classes.
Let $\mathbf{z} = (\mathbf{x}, \mathbf{y})$ be a data point with label $k$, i.e.~$y_k = 1,\ y_i = 0\ \forall\ i \neq k$.
The model $f_{\bm \theta}$ is a deep network with last layer's parameter $W \in \mathbb{R}^{d_y \times d_h}$, where $d_h$ is the number of hidden neurons.
Let $\mathbf{u} \in \mathbb{R}^{d_h}$ be the activation of the penultimate layer.
The output is computed as follow
\begin{align*}
\mathbf{a} &= W \mathbf{u} \\
\hat{\mathbf{y}} &= \delta(\mathbf{a})
\end{align*}
where $\delta$ is the softmax output function.
The derivative of the loss at $\mathbf{z}$ w.r.t.~$W$ is
\begin{align}
\frac{\partial \ell(\mathbf{z})}{\partial W} &= \nabla_{\mathbf{a}}\ell(\mathbf{z})\ \mathbf{u}^\top \\
&= \begin{bmatrix}
\nabla_{\mathbf{a}}\ell(\mathbf{z})_1 \mathbf{u}^\top \\
\vdots \\
\nabla_{\mathbf{a}}\ell(\mathbf{z})_{d_y} \mathbf{u}^\top
\end{bmatrix} \label{eqn:dw}
\end{align}
The gradient $\nabla_{\mathbf{a}}\ell(\mathbf{z})$ is
\begin{align}
(\nabla_{\mathbf{a}}\ell)^\top &= \frac{\partial \ell}{\partial \mathbf{a}} \\
&= \frac{\partial \ell}{\partial \hat{\mathbf{y}}} \frac{\partial \hat{\mathbf{y}}}{\partial \mathbf{a}} \\
&= \begin{bmatrix}
\frac{\partial \ell}{\partial \hat{y}_1} &
\cdots &
\frac{\partial \ell}{\partial \hat{y}_k} &
\cdots &
\frac{\partial \ell}{\partial \hat{y}_{d_y}}
\end{bmatrix} \times \nonumber \\
&\begin{bmatrix}
\frac{\partial \hat{y}_1}{\partial a_1} & \frac{\partial \hat{y}_1}{\partial a_2} & \cdots & \frac{\partial \hat{y}_1}{\partial a_{d_h}} \\
\vdots & \vdots & \vdots & \vdots \\
\frac{\partial \hat{y}_k}{\partial a_1} & \frac{\partial \hat{y}_k}{\partial a_2} & \cdots & \frac{\partial \hat{y}_k}{\partial a_{d_h}} \\
\vdots & \vdots & \vdots & \vdots \\
\frac{\partial \hat{y}_{d_y}}{\partial a_1} & \frac{\partial \hat{y}_{d_y}}{\partial a_2} & \cdots & \frac{\partial \hat{y}_{d_y}}{\partial a_{d_h}} \\
\end{bmatrix} \label{eqn:fullGrad} \\
&= \begin{bmatrix}
\frac{\partial \ell}{\partial \hat{y}_k} \frac{\partial \hat{y}_k}{\partial a_1} & \cdots & \frac{\partial \ell}{\partial \hat{y}_k} \frac{\partial \hat{y}_k}{\partial a_k} & \cdots & \frac{\partial \ell}{\partial \hat{y}_k} \frac{\partial \hat{y}_k}{\partial a_{d_h}}
\end{bmatrix} \label{eqn:reducedGrad}
\end{align}
We go from Eqn.~\ref{eqn:fullGrad} to Eqn.~\ref{eqn:reducedGrad} by using the following fact 

\begin{align*}
\frac{\partial \ell}{\partial \hat{y}_i} = \begin{cases}
0 \textnormal{ if } i \neq k \\
\frac{1}{\hat{y}_i} \textnormal{ if } i = k
\end{cases}
\end{align*}

We also have
\begin{align*}
\frac{\partial \hat{y}_k}{\partial a_i} = \begin{cases}
\hat{y}_k (1 - \hat{y}_k) \textnormal{ if } i = k \\
-\hat{y}_k \hat{y}_i \textnormal{ if } i \neq k
\end{cases}
\end{align*}
Substitute this into Eqn.~\ref{eqn:reducedGrad} we have
\begin{align*}
\nabla_{\mathbf{a}}\ell = \begin{bmatrix}
-\hat{y}_1 \\
\vdots \\
1 - \hat{y}_k \\
\vdots \\
-\hat{y}_{d_y}
\end{bmatrix}
\end{align*}
Because $1 - \hat{y}_k = \sum_{j \neq k} \hat{y}_j$, $1 - \hat{y}_k$ is much greater than $\hat{y}_j$ in general.
Substitute this into Eqn.~\ref{eqn:dw}, we see that the magnitude of the $k$-th row is much larger than than of other rows.
We also note that the update for the $k$-th row of $W$ has the opposite direction of the updates for other rows.

Let's consider the inner product of the gradients of two data points $\mathbf{z}$ and $\mathbf{z}'$ with label $k$ and $k'$. Let's consider the case where $k' \neq k$ first.

\begin{align}
\text{vec}\left( \frac{\partial \ell(\mathbf{z})}{\partial W} \right)^\top \text{vec}\left( \frac{\partial \ell(\mathbf{z}')}{\partial W} \right) = (\nabla_{\mathbf{a}}\ell^\top \nabla_{\mathbf{a}'}\ell) (\mathbf{u}^\top \mathbf{u}')
\end{align}

Intuitively, the product $\nabla_{\mathbf{a}}\ell^\top \nabla_{\mathbf{a}'}\ell$ is small because the large element $\nabla_{\mathbf{a}}\ell_k = 1 - \hat{y}_k$ is multiplied to the small element $\nabla_{\mathbf{a}'}\ell_k = \hat{y}'_k$ and the large element $\nabla_{\mathbf{a}'}\ell_{k'} = 1 - \hat{y}'_{k'}$ is multiplied to the small element $\nabla_{\mathbf{a}}\ell_{k'} = \hat{y}_{k'}$.
To make it more concrete, let's assume that $\hat{y}_k = \alpha \approx 1$ and $\hat{y}_i = \frac{1 - \alpha}{d_y - 1} = \beta$ for $i \neq k$.
We assume the same condition for $\hat{\mathbf{y}}'$.

\begin{align}
\nabla_{\mathbf{a}}\ell^\top \nabla_{\mathbf{a}'}\ell &= (\hat{y}_k - 1) \hat{y}'_k + (\hat{y}'_{k'} - 1)\hat{y}_{k'} + \sum_{i = 1, i \neq k, k'}^{d_y} \hat{y}_i \hat{y}'_i \nonumber \\
&= (d_y - 2)\beta^2 - 2(d_y - 1)\beta^2 \nonumber \\
&= -d_y \beta^2 \nonumber \\
&= -\frac{d_y(1 - \alpha)^2}{(d_y - 1)^2} \label{eqn:kNeqK}
\end{align}
$\alpha \approx 1$ implies $1 - \alpha \approx 0$ and $\beta \approx 0$.
Eqn.~\ref{eqn:kNeqK} implies that as the model is more confident about the label of $\mathbf{z}$ and $\mathbf{z}'$, the product $\nabla_{\mathbf{a}}\ell^\top \nabla_{\mathbf{a}'}\ell$ tends toward 0 at a quadratic rate.
The means, as the training progresses, data points from different classes become more and more independent.
The gradients of data points from different classes also become more and more perpendicular.

The sign of the gradient product depends on the sign of $\nabla_{\mathbf{a}}\ell^\top \nabla_{\mathbf{a}'}\ell$ and $\mathbf{u}^\top \mathbf{u}'$.
The signs of $\nabla_{\mathbf{a}}\ell^\top \nabla_{\mathbf{a}'}\ell$ and $\mathbf{u}^\top \mathbf{u}'$ are random variables that depend on the noise in the features $\mathbf{u}$ and $\mathbf{u}'$ and the weight matrix $W$.
If the model $f_{\bm \theta}$ cannot learn a good representation of the input then the feature $\mathbf{u}$ and the sign of $\mathbf{u}^\top \mathbf{u}'$ could be very noisy.
$\text{sign}(\mathbf{u}^\top \mathbf{u}')$ is even noisier if $\mathbf{z}$ and $\mathbf{z}'$ are from different classes.
Because $\abs{\nabla_{\mathbf{a}}\ell^\top \nabla_{\mathbf{a}'}\ell}$ is small, a tiny noise in the logits $\mathbf{a}$ and $\mathbf{a}'$ can flip the sign of $\nabla_{\mathbf{a}}\ell^\top \nabla_{\mathbf{a}'}\ell$ and change the direction of influence.

We now consider the case where $k' = k$. 
When $k' = k$, $\nabla_{\mathbf{a}}\ell^\top \nabla_{\mathbf{a}'}\ell$ is always positive. 
The sign of the gradient product only depends on $\mathbf{u}^\top \mathbf{u}'$.
That explains why the product of gradients of data points from the same class is much less noisy and almost always is positive.

Furthermore, the magnitude of $\nabla_{\mathbf{a}}\ell^\top \nabla_{\mathbf{a}'}\ell$ is larger than that in the case $k' \neq k$ because the large element $1 - \hat{y}_k$ is multiplied to the large element $1 - \hat{y}'_k$.
More concretely, under the same assumption as in the case $k' \neq k$, we have
\begin{align}
\nabla_{\mathbf{a}}\ell^\top \nabla_{\mathbf{a}'}\ell &= (1 - \hat{y}_k) (1 - \hat{y}'_{k}) + \sum_{i = 1, i \neq k}^{d_y} \hat{y}_i \hat{y}'_i \nonumber \\
&= (1 - \alpha)^2 + (d_y - 1)\beta^2 \label{eqn:kEqk}
\end{align}
From Eqn.~\ref{eqn:kEqk}, we see that when $k' = k$, the magnitude of $\nabla_{\mathbf{a}}\ell^\top \nabla_{\mathbf{a}'}\ell$ is approximately $d_y$ times larger than that when $k' \neq k$.

\end{document}